\pdfoutput=1
\documentclass[11pt]{article}

\usepackage{acl2023}

\usepackage{times}
\usepackage{latexsym}
\usepackage{amsmath}
\usepackage{mathtools}
\usepackage{amssymb}
\usepackage{bbm}
\usepackage{tikz}
\usepackage{tikz-qtree}
\usepackage{paralist, tabularx}
\usepackage[most]{tcolorbox}
\usepackage{enumitem}

\usepackage{algorithm}
\usepackage[noend]{algpseudocode}

\usepackage{booktabs} 
\usepackage{multicol}
\usepackage{multirow}
\usepackage{hyperref}

\usepackage{listings}

\usepackage{pifont}

\definecolor{Gray}{gray}{0.9}
\usepackage{soul}

\usepackage{color}

\newcommand{\counterfactual}{\ensuremath{
  \Box\kern-1.5pt
  \raise1pt\hbox{$\mathord{\rightarrow}$}}}

\newcommand{\disco}{$\mathcal{DISCO}$\xspace}

\makeatletter
\newcommand{\printfnsymbol}[1]{%
  \textsuperscript{\@fnsymbol{#1}}%
}
\makeatother

\usepackage[T1]{fontenc}

\usepackage[utf8]{inputenc}
\usepackage{microtype}
\usepackage{arydshln}

\usepackage{xspace}
\usepackage{listings} 
\title{$\mathcal{DISCO}$: Distilling Counterfactuals with Large Language Models}

\author{
  Zeming Chen\printfnsymbol{2}\thanks{ \hspace{5pt}  Work done while at the Allen Institute for AI. Equal contribution} \: Qiyue Gao\printfnsymbol{3}\textsuperscript{$*$} \: Antoine Bosselut\printfnsymbol{2} \: Ashish Sabharwal\printfnsymbol{3} \: Kyle Richardson\printfnsymbol{3} \\
  \printfnsymbol{2} Natural Language Processing Lab, EPFL, Lausanne, Switzerland \\
    { \normalsize \tt \{zeming.chen, antoine.bosselut\}@epfl.ch }
\\ 
  \printfnsymbol{3} Allen Institute for AI, Seattle, U.S.A. \\
  { \normalsize \tt \{bertg, kyler, ashishs\}@allenai.org}
}
    
\begin{document}
\maketitle

\begin{abstract}
Models trained with counterfactually augmented data learn representations of the causal structure of tasks, enabling robust generalization. However, high-quality counterfactual data is scarce for most tasks and not easily generated at scale. When crowdsourced, such data is typically limited in scale and diversity; when generated using supervised methods, it is computationally expensive to extend to new counterfactual dimensions. In this work, we introduce \disco  (\textbf{DIS}tilled \textbf{CO}unterfactual Data), a new method for automatically generating high-quality counterfactual data at scale. \disco engineers prompts to generate phrasal perturbations with a large general language model. Then, a task-specific teacher model filters these generations to distill high-quality counterfactual data. While task-agnostic, we apply our pipeline to the task of natural language inference (NLI) and find that on challenging evaluations such as the NLI stress test, comparatively smaller student models trained with \disco-generated counterfactuals are more robust (6\% absolute) and generalize better across distributions (2\%) compared to models trained without data augmentation. Furthermore, \disco-augmented models are 10\% more consistent between counterfactual pairs on three evaluation sets, demonstrating that \disco augmentation enables models to more reliably learn causal representations. Our repository is available at: \href{https://github.com/eric11eca/disco}{https://github.com/eric11eca/disco}
\end{abstract}

\section{Introduction}

\begin{figure}
\begin{center}
\includegraphics[width=0.95\columnwidth]{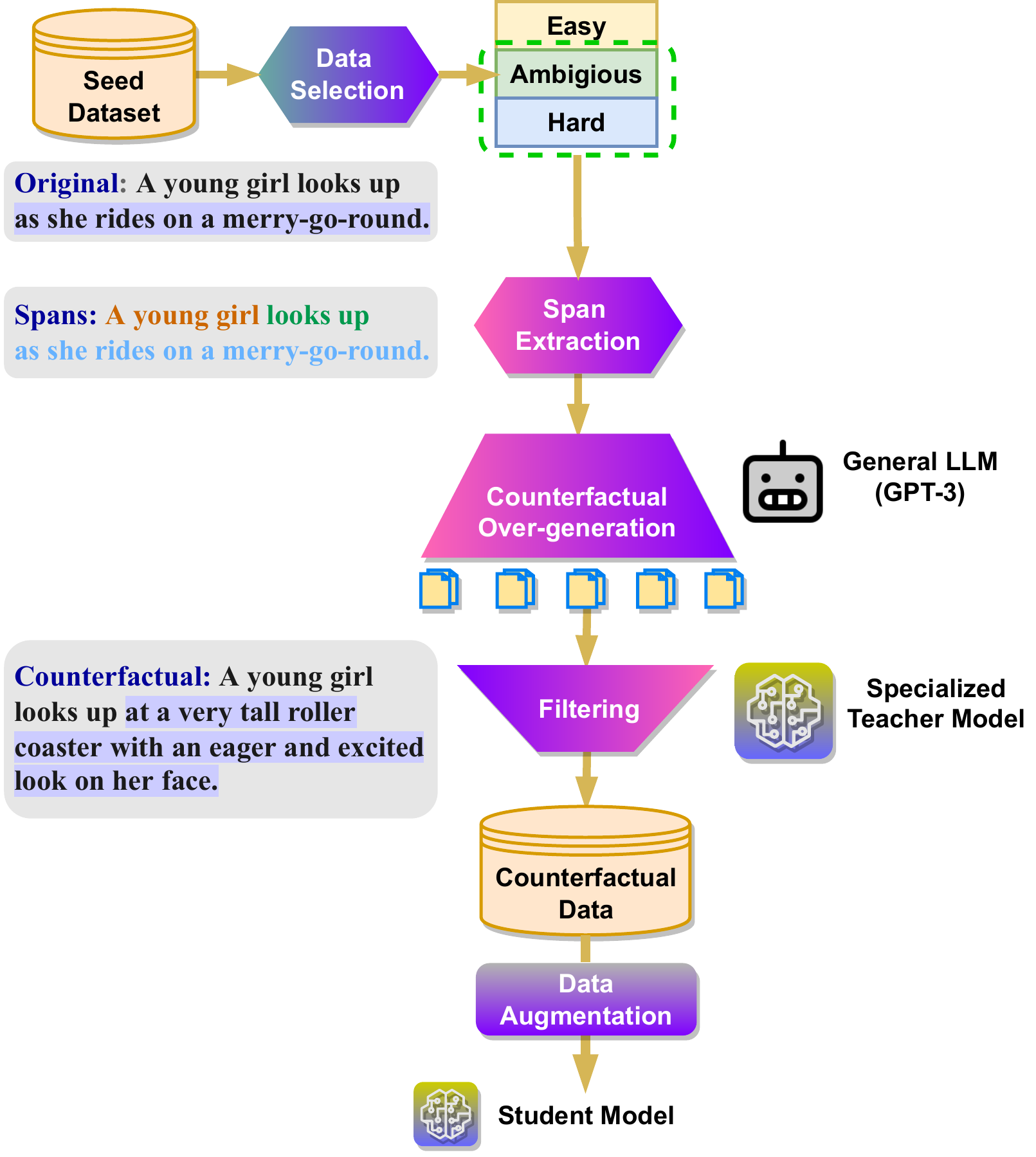}
\end{center}
\caption{Overview of our counterfactual data distillation process (\disco) using a large language model.}
\label{fig:splash}
\end{figure}

Despite the tremendous progress made in NLP on a wide range of reasoning tasks \cite{glue2018,wang2019superglue,clue}, dataset biases continue to be a formidable challenge for robust model development \cite{gururangan2018annotation,poliak2018hypothesis,kaushik2018much,tsuchiya2018performance,liu-2020-empirical,shortcut2022}. Counterfactual data augmentation (CAD) \cite{kaushik2019learning} is one general approach to improve model robustness by training on edited instances that systematically alter the critical or causally salient parts of dataset instances that contributes to the label assignment. To date, two main approaches have been pursued as part of these efforts: \emph{human-centered approaches}, where edits are obtained through direct human annotation and crowdsourcing \cite{kaushik2019learning,khashabi2020more,gardner2020evaluating}; and \emph{model-based approaches}, where new examples are collected through automatic text generation \cite[][\emph{inter alia}]{wu2021polyjuice,madaan2021generate,ross2021tailor,wen2022autocad}.  

However, crowd-sourcing counterfactual data can be inefficient, costly, and difficult to scale. This often results in small counterfactual datasets, which can hinder the diversity and coverage of the collected edits (e.g., in \citet{kaushik2019learning}, the training scenario for NLI involves 8.3k total instances with augmentation). In contrast, supervised text generation methods are cheaper and easier to scale (e.g., \citet{wu2022generating} use generation methods that expand NLP datasets to include around a million total examples). However, such methods can only generate fixed perturbation types. They rely on a fixed inventory of perturbation types each requiring new training sets. This is hard to scale up and can limit the space of perturbation types learned by the corresponding learned generation models. They can also be expensive to extend to new perturbation types, given the need to retrain models.

In this paper, we focus on the Natural Language Inference (NLI) task, which has recently been shown to benefit from collaboration between human annotation and LLMs in the WANLI data augmentation system of \citet{liu2022wanli}. Our primary contribution is a \emph{counterfactual knowledge distillation} procedure called \disco (\textbf{DIS}tilled \textbf{CO}unterfactual Data), which works in the following way (see Figure~\ref{fig:splash}): First, task instances to be edited are selected and decomposed into spans using off-the-shelf linguistic processing tools. Then prompt engineering and in-context learning are applied with a general LLM to overgenerate a diverse set of perturbations for these instances. We then employ a large \emph{teacher} NLI model to conservatively filter the over-generations as a fully-automatic alternative to the human filtering used in WANLI. The distilled generations are finally used to train a much smaller and high-performance \emph{student model}.

We show that \disco, despite not relying on explicit human annotation, yields high-quality datasets. Manual annotation shows that, on average, 83\% of our counterfactual data correctly flips the source labels, which is 1\% higher than human performance. Additionally, compared to human CAD examples \cite{kaushik2019learning}, we find \disco generated data to have much-improved perturbation and information richness. Through data augmentation experiments, we also find that training on datasets built using \disco obtains competitive and often improved performance across a wide range of robustness and out-of-domain (OOD) NLI tests, despite having a significantly smaller size than existing augmentation approaches (75k vs. 1 million from \citet{wu2022generating}). This includes consistent improvements (6\% average) over WANLI and SNLI baselines on 7 NLI robustness tests. Building on the impressive results from \citet{liu2022wanli}, this is significant as it shows the promising potential of data augmentation via LLMs, even without explicit human annotation. We find that models trained using our data exhibit 8\% improved counterfactual accuracy and 6\% increased sensitivity to context differences between counterfactual pairs than SNLI baselines. When augmenting on top of WANLI, our method shows an 18\% performance gain on counterfactual accuracy.

\paragraph{Contributions} In summary, we present \disco, a fully-automatic counterfactual knowledge distillation approach based on LLMs. To our knowledge, \disco is the first to use LLMs such as GPT3 for counterfactual data augmentation. We show that our approach helps produce more diverse counterfactuals over existing crowd-sourcing approaches while showing higher quality than human-written data. The distilled counterfactual data is more effective than existing augmentation approaches for improving NLI robustness, OOD generalization, and counterfactual consistency.

\section{Related Work}

\paragraph{Mitigating Spurious Correlations for NLU}

The augmentation methods described above are part of a large literature on model debiasing approaches, which also includes work on dataset filtering \cite{AFLite}, model ensembling \cite{clark2019don}, feature removal, and other learning-based techniques \cite{belinkov2019don,mahabadi2019end}. \citet{wu2022generating} also propose a new debiasing method called Z-Aug that learns to generate unbiased samples and filter out biased data using a z-statistic filter. In contrast to the debiasing and data generation techniques already discussed, our approach is unique in exploiting the power of LLMs such as GPT3 \cite{brown2020language} to create more diverse augmented datasets as a way to mitigate biases and shortcuts. 

\paragraph{Counterfactual Data Augmentation}
Augmenting models with counterfactual data is a popular recent approach for mitigating spurious correlation and improving model robustness. \citet{kaushik2019learning} first recruits human workers to write counterfactual examples for augmentation. They find that counterfactually augmented data can help mitigate spurious patterns in the training data. As already discussed, however, creating counterfactual data using humans requires a high cost, is time-consuming, and can result in simple perturbations. Later, \citet{wu2021polyjuice} and \citet{ross2021tailor} proposed frameworks that use text generation models to generate counterfactual data. These models require fine-tuning using pre-defined perturbation types. Both methods have constraints: (1) the generation is un-targeted, thus unlabeled, and (2) the perturbation types are limited. To acquire new perturbation types, the models have to be re-trained. Unlike the previous methods, our method uses LLMs to generate more diverse perturbation types cheaply and efficiently. Our method also improves over un-targeted generations by using a task-specific teacher model to verify the label.

\paragraph{Large Model Dataset Creation}
Leveraging the powerful generative ability of large language models to create datasets automatically has recently attracted considerable attention. This method reduces the cost of manually creating the dataset, can collect more diverse phenomena to expand the distribution, and can be adapted to a wide range of tasks in NLP. The most similar work to ours is WANLI \cite{liu2022wanli}, an NLI dataset fully generated by GPT-3 and annotated by human workers. The idea is to elicit ambiguous NLI examples from GPT-3  to improve its performance on challenge evaluation benchmarks, which relies on the \emph{dataset cartography} techniques from \citet{swayamdipta2020dataset} that we also use in our study for selecting instances to edit. Our work also seeks to get diverse data from GPT-3 to improve model robustness. Differently, we only make local perturbations on the premise instead of generating a new example. We did not label our training data using human workers but leveraged an NLI model to filter out the counterfactual examples.

\section{Counterfactual Distillation}
The central idea of counterfactual data distillation is to prompt a large language model through in-context learning to generate perturbations that can flip the current label to a new one (ex. $Contradiction \rightarrow Entailment$). Once we select a subset of a dataset (discussed in Section \ref{sec:da}), we first identify potential locations for performing counterfactual perturbations on the target instances. Then we prompt the GPT-3 (text-DaVinci-002) model to overgenerate perturbations (\ref{sec:gpt3_gen}). We use a teacher language model specializing in the NLI task to filter the generated perturbations based on the shift in model predictions from the original to the new label (\ref{sec:filter}). Formally, given an input premise-hypothesis pair $<\mathrm{P}, \mathrm{H}>, l$ where $l \in \{Entailment, Contradiction, Neutral\}$ is the ground-truth label. We want to get a counterfactual input $<\mathrm{P}', \mathrm{H}>, l'$ where we get $\mathrm{P}'$ by perturbing parts of the premise $\mathrm{P}$ and $l'$ is the new label corresponding to the new input. 

\begin{figure*}[t]
\begin{center}
\includegraphics[width=0.9\textwidth]{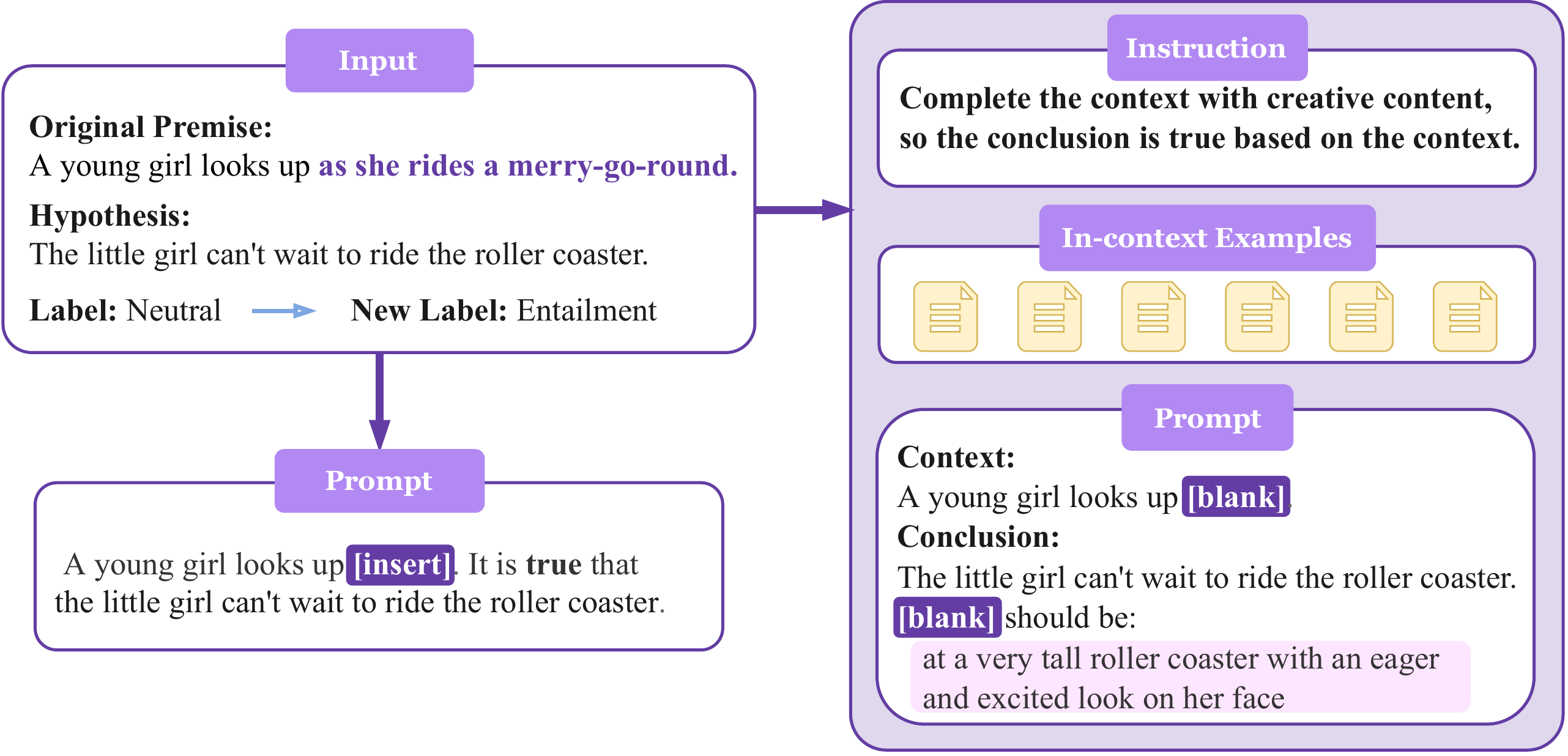}
\end{center}
\caption{Overview of the perturbation generation process on GPT-3 using the masked NLI prompt (right) and the insertion NLI prompt (left-bottom). Here we are editing the input premise and hypothesis for the given NLI problem to change the label from \emph{Neutral} to \emph{Entailment}. Thus we have "conclusion is true`` in both prompts.}
\label{fig:masked}
\end{figure*}

\subsection{Prompting} \label{sec:gpt3_gen}

We experiment with various prompting strategies on GPT-3, detailed and illustrated in Figure~\ref{fig:masked}. To make local edits to a sentence following CAD \cite{kaushik2019learning}'s procedure, we use a neural syntactic parser \cite{parser} to split sentences to perturb into spans. Using this neural chunker, we can get a set of spans $\mathcal{S} = \{s: s \in \mathrm{P}\}$ decomposed from the premise $\mathrm{P}$. These spans serve as the potential locations for making a perturbation.

\paragraph{Masked Prompting.}
To prompt GPT-3 for counterfactual perturbations, we use a masked NLI format to build the prompt. Let $\mathrm{P}$ and $\mathrm{H}$ be the premise and hypothesis pair we want to perturb, associated with the current label $l$ and the set of spans $\mathcal{S}$.  We select one span from $\mathcal{S}$ and replace it in the premise with a mask token \textbf{[blank]}. Given a new label $l'$ we want to flip to, we ask the model to fill in the blank mask token with creative perturbation $s'$ to get a new premise $\mathrm{P}'$ that satisfies $l'$. Here the perturbation serves as an intervention in flipping the original label to the new label. Because during the generation time, one can not know which span will flip the label after perturbation, we overgenerate perturbations by iterating through all the spans from a premise. Each span yields a new prompt and makes a new request to GPT-3. 

\paragraph{Insertion Mode.}
One of the key features of GPT-3 is its insertion mode, which allows users to insert a piece of text into the current context and have the model generate text based on the surrounding context. We can naturally convert the masked-NLI prompt into an insertion prompt format by providing the surrounding text of the mask token to the model. By forming a natural sentence, we try to align the prompt to the pre-training objective of GPT-3 (e.g., casual language modeling). We first map the label space $\{Entailment, Contradiction, Neutral\}$ to $\{true, false, possible\}$. Then we build the prompt: "<Prefix> [insert] <Suffix>. It is <$l'$> that <$\mathrm{H}$>", where $l'$ is the new label.

The advantage of using the insertion mode is that the model considers both the prefix and suffix context of the masked span. This solves a common issue in the completion mode where the model tends to finish a sentence when generating the perturbation without noticing the suffix context. Additionally, the insertion mode does not require in-context learning examples, which yields more diverse generations at a much lower cost.

\subsection{Teacher Model Filtering} \label{sec:filter}

Using a combination of the prompting strategies detailed in the last section, we then implement a filtering system to select the most promising counterfactual examples, pruning out potential mistakes made by GPT3. The filtering system first uses a heuristic-based automatic filter to remove any generations that yield obvious signs of low quality, ensuring that the remaining perturbations are more likely to flip the label in the desired direction. Our check for several criteria, including:
\begin{enumerate}[leftmargin=*,nosep]
    \item Does the perturbation contain parts from the instruction or prompt?
    \item Does the perturbation copy parts from the in-context examples?
    \item Does the perturbation repeat parts from the premise or hypothesis?
\end{enumerate} Using a count of the lexical overlap rate between sentences and a pre-defined set of common negation words, we also remove any perturbations with clear data artifacts, such as excessive lexical overlap between premise and hypothesis or using negation words as a shortcut to flip the label. After the automatic filtering, we distill the remaining data using a model-based teacher, which identifies the perturbations that convert the original label to the target label. To verify if a perturbation potentially converts the original label in the direction of the new label, a natural way would be to check if the prediction probability of the new label shifts by a large margin, given the new input and the original input. Specifically, we calculate the distributional shift as follows: \\
\begin{equation}
    \Delta_{l'} = p(l' | \mathrm{P}',\mathrm{H}) - p(l' | \mathrm{P}, \mathrm{H})\text{,} \\ 
\end{equation}
which yields the change in prediction probability from the original input to the new input. We use a DeBERTa-v2 \cite{deberta} model with SOTA performance on NLI as the teacher model. Additional details about the prompting parameters and teacher model can be found in Appendix \ref{sec:hypers}.

\section{Evaluate Counterfactual Quality}
Large general language models like GPT-3 enable the generation of counterfactual data at a large scale. The generation process is more efficient, cheap, and flexible than crowdsourcing. Here we evaluate the quality and diversity of \disco data against counterfactually augmented data written by human workers (Human-CAD) \cite{kaushik2019learning} using automatic and human-based metrics.

\subsection{Automatic Evaluation}
\paragraph{Diversity Measurement} Following other work on CAD \cite{wu2021polyjuice}, we use Self-BLEU \cite{selfbleu} to measure the \emph{diversity} of the generated counterfactual examples.  In Table \ref{tab:evaluation}, we list the Self-BLEU score for each perturbation direction. Compared to human-written examples, GPT-3 generated examples have much lower Self-BLEU scores than human-written ones indicating that GPT-3 can generate far more diverse examples. 

\paragraph{Dataset Distance} The Self-BLEU score measures lexical and syntactic diversity only. To assess the diversity of information in the data, we calculate the dataset distance between the original examples and the new examples. Specifically, we measure dataset distance via OTDD (\emph{optimal transport dataset distance}) \cite{otdd-nips}, a model-agnostic distance metric that can operate on datasets with disjoint label sets. OTDD can measure how well the knowledge from one dataset can transfer to another. We use OTDD to assess the distributional difference between the original and new examples. As Table \ref{tab:evaluation} shows, our generated examples have a higher distance from the original examples than the human-written data, consistently in all directions. This trend demonstrates that our counterfactual data provide more diverse information than human-written data. 

\begin{table}[!t]
\begin{center}
\resizebox{\columnwidth}{!}{
    \begin{tabular}{lccccccc}
      \toprule  
       & \multicolumn{7}{c}{\bf Flip Rate Score $\uparrow$} \\
       Data &  \multicolumn{1}{c}{\bf{E2C}} & \multicolumn{1}{c}{\bf{E2N}} & \multicolumn{1}{c}{\bf{N2C}} & \multicolumn{1}{c}{\bf{N2E}} & \multicolumn{1}{c}{\bf{C2N}} & \multicolumn{1}{c}{\bf{C2E}} & \multicolumn{1}{c}{\bf{Avg.}}\\ \midrule
       {\bf Human-CAD} & 86.37 & 82.36 & 86.08 & 84.34 & 73.42 & 82.28 & 82.55 \\ 
       {\bf $\mathcal{DISCO}$ (ours)} & 78.53 & 82.70 & 76.20 & 85.53 & 75.76 & 92.43 & 83.14 \\ \midrule

       & \multicolumn{7}{c}{\bf Soft Flip Rate Score $\uparrow$} \\  
       & \multicolumn{1}{c}{\bf{E2C}} & \multicolumn{1}{c}{\bf{E2N}} & \multicolumn{1}{c}{\bf{N2C}} & \multicolumn{1}{c}{\bf{N2E}} & \multicolumn{1}{c}{\bf{C2N}} & \multicolumn{1}{c}{\bf{C2E}} & \multicolumn{1}{c}{\bf{Avg.}}\\
       \midrule
       {\bf Human-CAD} & 94.32 & 83.33 & 88.61 & 86.75 & 82.28 & 94.94 & 88.24 \\ 
       {\bf $\mathcal{DISCO}$ (ours)} & 97.55 & 88.46 & 76.20 & 89.47 & 92.42 & 95.45 & 93.33\\ \midrule

      & \multicolumn{7}{c}{\bf Self-BLEU Diversity Score $\downarrow$} \\
      & \multicolumn{1}{c}{\bf{E2C}} & \multicolumn{1}{c}{\bf{E2N}} & \multicolumn{1}{c}{\bf{N2C}} & \multicolumn{1}{c}{\bf{N2E}} & \multicolumn{1}{c}{\bf{C2N}} & \multicolumn{1}{c}{\bf{C2E}} & \multicolumn{1}{c}{\bf{Avg.}}\\
      \midrule
      {\bf Human-CAD} & 0.76 & 0.75 & 0.82 & 0.82 & 0.81 & 0.79 & 0.79 \\ 
      {\bf $\mathcal{DISCO}$ (ours)} & 0.23 & 0.26 & 0.26 & 0.18 & 0.25 & 0.21 & 0.23 \\ \midrule

      & \multicolumn{7}{c}{\bf OTDD Dataset Distance $\uparrow$} \\
      & \multicolumn{1}{c}{\bf{E2C}} & \multicolumn{1}{c}{\bf{E2N}} & \multicolumn{1}{c}{\bf{N2C}} & \multicolumn{1}{c}{\bf{N2E}} & \multicolumn{1}{c}{\bf{C2N}} & \multicolumn{1}{c}{\bf{C2E}} & \multicolumn{1}{c}{\bf{Avg.}}\\
      \midrule
      {\bf Human-CAD} & 217 & 95 & 179 & 139 & 238 & 217 & 180 \\ 
      {\bf $\mathcal{DISCO}$ (ours)} & 250 & 199 & 254 & 165 & 275 & 301 & 240 \\
      \bottomrule
    \end{tabular}
}
\caption{Automatic and human evaluation results on a random subset (510 instances) of our counterfactual data ($\mathcal{DISCO}$), compared with Human-CAD \cite{kaushik2019learning}, counterfactual data written by human workers.
} \label{tab:evaluation}
\end{center}
\end{table}

\subsection{Human Evaluation}
\paragraph{Label-Flip Score}
The label-flip score is an accuracy-based metric to check if the new example after perturbation forms a counterfactual to the original example. We check the flip score in two aspects. The Label Flip Rate (LFR) calculates the percentage of new examples that flip the original label to the target label. The Soft Label Flip Rate (SLFR) calculates the percentage of new examples whose label differs from the original example's label. SLFR measures how often LLMs generate valid counterfactuals independent of whether the new label is right. Given the rigidness of LFR and the fluidity of some NLI judgements \cite{pavlick2019inherent}, this last metric is meaningful for checking if we still generate valid counterfactuals even when the exact label is not correct. The high SLFR suggests that many examples not accepted by the filter could be valid counterfactuals making them useful for other types of learning (e.g., leveraging signals from such data to train models to identify counterfactuals). For a dataset with $K$ examples, we calculate FLR and SFLR as follows:
\begin{align*}
    \mathrm{LFR} &= \frac{1}{K} \sum^{K}_{k=1} \mathbbm{1}(\Tilde{l}_k = l'_k) \\
    \mathrm{SLFR} &= \frac{1}{K} \sum^{K}_{k=1} \mathbbm{1}(\Tilde{l}_k \neq l_k)\text{,}
\end{align*}
where $\Tilde{l}$ is the annotated label, $l'$ is the target label, and $l$ is the original label.

We use Amazon Mechanic Turk to conduct human evaluations, asking annotators to label a random subset of our data following the standard annotation process for the NLI task. We assigned three annotators for each example and did majority voting on the annotated labels. We list more details on the instructions, interface, and annotator requirements in Appendix \ref{sec:amt}. We only give annotators the new sentence pairs to avoid bias from the original example. Table \ref{tab:evaluation} shows the human evaluation results in each perturbation direction. 

Compared to human-written examples, \disco has lower LFRs only on generating contradictions, showing that GPT-3 generates better entailment and neutral examples rather than contradiction examples. We hypothesize that this is due to the ambiguous boundary between contradiction and neutral examples. Moreover, generating contradictions while maintaining diversity is difficult. When asked to generate contradictions, they tend to generate neutral examples by changing a sentence's semantics (i.e., adding diversified words). In the case of Human-CAD, annotators tend to create contradictions using simple tricks like negation \cite{joshi-he-2022-investigation}. Although these tricks can produce absolute contradiction examples, they can introduce strong data artifacts, leading to a model that is not robust.
Overall, the human evaluation scores show that our distilled counterfactual data exceeds human-written examples in correctly flipping the label, as shown by a higher average flip rate score.

\section{Experiments} \label{sec:exp}

\begin{table}[!t]
    \centering
    \resizebox{\columnwidth}{!}{
    \begin{tabular}{llcc}
         \toprule
         \bf{Dataset} & \bf{Focus} & \bf{Size} \\ \hline
         PI-CD (a) & Partial-input heuristics & 3261  \\
         PI-SP (b) & Partial-input heuristics & 371  \\
         IS-CS (c) & Inter-sentences Heuristics & 656  \\
         LI-LI (d,e) & Logical Inference Ability  & 9927  \\
         LI-TS (f,g) & Logical Inference Ability  & 9832  \\
         ST (e) & Stress (distraction \& noise) test & 93447  \\ 
         HANS (h) & Syntactic Heuristic & 30000 \\ \hline
         MNLI-hard-m & Out-of-distribution &  4573 \\
         MNLI-hard-mm & Out-of-distribution &  4530 \\
         QNLI  & Out-of-distribution &  5266 \\ \hline
         Human-CAD & Counterfactual consistency & 1600  \\
         SNLI-hard$_{\counterfactual}$ & Counterfactual consistency & 3042  \\
         WANLI$_{\counterfactual}$ & Counterfactual consistency & 4000  \\
         \bottomrule
    \end{tabular}}
    \\
    \resizebox{\columnwidth}{!}{\begin{tabular}{ll}
        (a) \citet{gururangan2018annotation} & (b) \citet{liu-hyponli} \\
        (c) \citet{Nie_Wang_Bansal_2019} & (d) \citet{glockner2018breaking} \\
        (e) \citet{naik2018stress} & (f) \citet{minervini2018adversarially} \\
        (g) \citet{whatifwang2018} & (h) \citet{mccoy2019right} \\
    \end{tabular}}
    \caption{Details about the evaluation datasets we used for the experiments.}
    \label{tab:comparison}
\end{table}

\begin{table*}[t!]
\begin{center}
\resizebox{\textwidth}{!}{
    \begin{tabular}{lccccccccccc}
        \toprule
        & & \multicolumn{7}{c}{\bf{Model Robustness}} & \multicolumn{3}{c}{\bf{OOD Generalization}} \\
        \cmidrule(lr){3-9}  \cmidrule(lr){10-12}
        {\bf Method}  & \bf{Size} & \textbf{PI-CD} & \textbf{PI-SP} & \textbf{IS-CS} & \textbf{LI-LI} & \textbf{LI-TS} & \textbf{ST} & \textbf{HANS} & \multicolumn{1}{c}{\bf{MNLI$^1$}} &  \multicolumn{1}{c}{\bf{MNLI$^2$}} &  \multicolumn{1}{c}{\bf{QNLI}} \\
        
        \midrule 
        \multicolumn{2}{l}{\bf{Large-size augmentation on full SNLI}} \\
        {SNLI} & 549,367 & 82.2 & 69.0 & 68.4 & 93.6 & 72.5 & 72.4 & 73.1 & 78.5 & 78.2 & 64.5 \\
        + WANLI & 652, 252 & 83.4 & \underline{82.7} & 69.5 & 86.2 & 84.3 & 67.4 & \underline{87.4} & 78.2 & 78.0 & \underline{78.6}  \\
        + Z-aug & \textbf{1,142,475} & \underline{\bf{84.1}} & 72.5 & \underline{72.6} & \underline{\bf{93.9}} & \underline{87.1} & \underline{75.4} & 68.3 & \underline{80.0} & \underline{\bf{80.7}} & 75.0 \\
        
        \midrule
        \multicolumn{2}{l}{\bf{Augmentation on subset of SNLI}} \\
        {SNLI-subset} & 100,000 & 82.0 & 71.7 & 65.1 & 85.5 & 83.9 & 69.5 & 65.8 & 78.0 & 79.1 & 73.4  \\
        + Tailor & 192,457 & 79.5 & 52.0 & 55.8 & 84.6 & 80.1 & 62.7 & 55.8 & 64.1 & 65.7 & 71.4 \\
        + Human-CAD & 108,330 & 82.8 & \underline{77.8} & 69.2 & 90.7 & 87.1 & 71.3 & 65.5 & 79.0 & 79.0 & 72.8 \\
        + $\mathcal{DISCO}$ (ours) & 165,418 & \underline{\bf{84.1}} & 74.1 & \underline{\bf{73.5}} & \underline{92.1} & \underline{88.4} & \underline{\bf{77.0}} & \underline{70.1} & \underline{\bf{80.5}} & \underline{80.2} & \underline{77.7} \\

        \midrule
        \multicolumn{2}{l}{\bf{Augmentation on WANLI}}  \\
        WANLI & 102,885 & 65.6 & 81.3 & 65.9 & 65.6 & 82.7 & 56.5 & \underline{\bf{89.4}} & 76.1 & 76.3 & 81.1 \\
        + \disco (ours) & 177,885 & \underline{82.8} & \underline{\bf{83.8}} & \underline{72.0} & \underline{86.8} & \underline{85.1} & \underline{68.6} & 87.4 & \underline{80.0} & \underline{78.7} & \underline{\bf{81.4}} \\

        \midrule
        \multicolumn{2}{l}{\bf{Trained on \disco (ours) data only}}  \\
        \disco (ours) & \textbf{75,000} & 83.5 & 77.4 & 73.3 & 89.4 & \bf{88.9} & 76.3 & 70.7 & 79.2 & 79.5 & 79.1 \\
      
        \bottomrule
    \end{tabular}
 }
\caption{Results on Stress-tests, robust NLI test suites \cite{liu-2020-empirical}, MNLI-hard, and QNLI. The bold numbers are the highest accuracy within a column, and the underlined numbers are the highest accuracy for each section. MNLI$^1$ refers to MNLI-hard-match, and MNLI$^2$ refers to MNLI-hard-mismatch.}
\label{tab:aug-results}
\end{center}
\end{table*}

\subsection{Counterfactual Data Augmentation} \label{sec:da}
We next investigate how distilled GPT-3 counterfactual data can improve model robustness and generalizability through data augmentation. Given a set of original data $\mathcal{D} = \{\mathcal{X}, \mathcal{Y}\}$, we generate a perturbation $\mathrm{z}$ for each example in a subset of $\mathcal{D}$ ($\mathcal{D}^s = \{\mathcal{X}^s, \mathcal{Y}^s\}$), and convert the original one to a counterfactual example: $\mathcal{D}^c = \{(x^c = z(x), y') | x \in \mathcal{X}^s, y \in \mathcal{Y}^s\}$. Next, we augment this subset by merging it with the counterfactual examples: $\mathcal{D}^a = \mathcal{D}^s \cup \mathcal{D}^c$. For additional data augmentation, we also select a base set $\mathcal{D}^b$ (a random subset from $\mathcal{D}$), merge it with the augmentation set $\mathcal{D}^a$ and remove any duplicated examples: $\mathcal{D}_{train} = \mathcal{D}^b \cup \mathcal{D}^a - \mathcal{D}^d$. We use models trained on base sets $\mathcal{D}^b$ alone as baselines and evaluate whether augmenting the base sets using DISCO data would improve the baselines’ performances following Z-aug \cite{wu2022generating} and WANLI \cite{liu2022wanli}. We train a smaller student model, based on  \textbf{RoBERTa-large} (355 million parameters) using the implementation from \citet{wolf2020transformers}, on $\mathcal{D}_{train}$ and $\mathcal{D}^a$. Then, we evaluate the model on a set of test datasets for measuring robustness and OOD generalizability. 

\paragraph{Source Datasets}
We select SNLI \cite{bowman2015large} as the source dataset for generating \disco data and for data augmentation. SNLI is a widely-used NLI dataset employed in numerous research studies. We apply data cartography \cite{swayamdipta2020dataset} to select the ambiguous part of SNLI. The paper suggests that training on ambiguous data yields more robust models. Our intuition is that enhancing the ambiguous set with counterfactual examples would benefit the model's learning. We also augment \disco on WANLI \cite{liu2022wanli} to analyze the benefits of counterfactual data augmentation on a dataset constructed via human-GPT-3 collaboration. 

\paragraph{Evaluation Datasets}
We first evaluate how robust model performance is under adversarial and stress tests. We select the adversarial datasets from \citet{liu-2020-empirical}'s benchmark for debiasing strategies and NLI stress test suite from \citealp{naik2018stress}'s work. Next, we evaluate the model's generalizability across different distributions. We select two datasets with a different distribution from the SNLI dataset: MNLI-hard (matched and mismatched) \cite{mahabadi2019end}, and QNLI \cite{glue2018}, a dataset adapted from the Stanford Question Answering Dataset \cite{rajpurkar2016squad}. Details about the evaluation datasets are included in Table \ref{tab:comparison}. 

\paragraph{Comparisons}
For naive comparison, we evaluate our models against baselines trained $\mathcal{D}^b$ only without data augmentation. Then, we compare our models to prior augmentation methods, including Tailor \cite{ross2021tailor}, WANLI \cite{liu2022wanli}, Z-aug \cite{wu2022generating}, and Human-CAD \cite{kaushik2019learning}. For WANLI and Z-aug, we also augment them on the full SNLI training set because of their large dataset sizes. In addition, we fine-tune a model only on \disco to compare with all the models above (see Appendix~\ref{sec:hypers} for more details about training and hyper-parameters).

\paragraph{Results} Table \ref{tab:aug-results} shows that our counterfactual data augmentation significantly improves over the baseline performance on most robustness datasets when augmenting the \disco dataset on a subset of SNLI. Augmenting or training with \disco data achieves the highest accuracy on 7 evaluation sets. When augmenting on WANLI, the augmented model achieved better average performance (75.1) on robustness than the baseline WANLI model (65.9). We list the average performance gain for robustness and OOD generalization in Table \ref{tab:gain}. We can see that \disco-augmented models improve model robustness over baselines by a large margin (6.5 SNLI and 9.5 WANLI).  These results show the efficacy of our counterfactual data in helping models mitigate multiple types of NLI data bias altogether. On out-of-distribution (OOD) generalization, models trained on \disco augmented data achieve a positive performance gain of 2.7 \% over the SNLI subset baseline and 2.1\% over the WANLI baseline. This suggests that augmenting with \disco helps the model generalize to datasets with distributional shifts. Compared to prior data augmentation methods, \disco data can more significantly improve model performance, showing that our method yields high-quality and effective augmentation data.

In addition, \disco is much smaller than other augmentation data like WANLI and Z-aug. Interestingly, training on \disco data yields better performance than these models trained on large datasets (on 7 datasets). 

\begin{table}[t!]
\begin{center}
\resizebox{0.9\columnwidth}{!}{
    \begin{tabular}{llccc}
      \toprule
       & \textbf{Test Metrics} & \textbf{Original} & \textbf{Augmented} & \textbf{$\Delta$} \\
      \midrule
      \multirow{3}{*}{\rotatebox[origin=c]{90}{\footnotesize SNLI-SUB}} & 
      Robustness Avg. & 71.0 & \underline{\textbf{77.5}} & 6.5 \\
      & OOD Avg. & 76.7 & \underline{\textbf{79.4}} & 2.7  \\
      & Acc$_{\counterfactual}$ Avg. & 47.1 & \underline{\textbf{55.2}} & 8.1 \\
      & $\delta_{s}$ Avg. & 58.6 & \underline{\textbf{64.9}} & 6.3 \\
      \midrule
      \multirow{ 3}{*}{\rotatebox[origin=c]{90}{\footnotesize WANLI}} & 
      Robustness Avg. & 65.9 & \underline{\textbf{75.1}} & 9.2 \\
      & OOD Avg. & 78.0 & \underline{\bf{80.1}} & 2.1 \\
      & Acc$_{\counterfactual}$ Avg. & 34.6 & \underline{\textbf{52.7}} & 18.1 \\
      & $\delta_{s}$ Avg. & 44.9 & \underline{\textbf{57.6}} & 12.7 \\
      \bottomrule
    \end{tabular}
}
\caption{Performance gain of data augmentation using \disco from baselines \emph{without} augmentation (i.e., using the base sets in the first column).}
\label{tab:gain}
\end{center}
\end{table}

\subsection{Counterfactual Evaluation}
In our second experiment, we investigate how \disco data can enhance counterfactual reasoning ability of models on NLI problems. Counterfactual reasoning is the ability to predict
how an alternative context, contrary to the present context, might have resulted in
different outcomes \cite{qin-counterfactual}. In the setting of NLI, we alter the current context with text perturbations sufficient to change the current label to a different one while spuriously correlated features remain identical. A model that relies heavily on spurious features will likely fail to predict both the original and counterfactual examples correctly \cite{feder-etal-2022-causal}. 

\paragraph{Evaluation Datasets} We first create two counterfactual evaluation datasets using GPT-3 to generate the perturbations. We recruit human workers on Amazon Mechanic Turk to annotate labels for the two datasets. \textbf{SNLI-hard$_{\counterfactual}$} is constructed using a subset of the SNLI-hard \cite{gururangan2018annotation} dataset. We pair each original example with the generated counterfactual example, where human annotators provide the gold label. In addition, we want to construct a dataset different from \disco's distribution. Thus, we select a subset from the WANLI test set and follow the same procedure as SNLI-hard$_{\counterfactual}$ to get a counterfactual evaluation set \textbf{WANLI$_{\counterfactual}$}. We assign three human workers to each problem to annotate the label. We list more details on the instructions, interface, and annotator requirements in Appendix \ref{sec:amt}. We also include the \textbf{Human-CAD} dataset as the examples were written and labeled by human workers.

\paragraph{Metrics}
\normalsize
We measure models' counterfactual reasoning ability along two dimensions. First, we measure \emph{counterfactual sensitivity} $\delta_{s}$: how confidently a model differentiates the original and counterfactual examples. In other words, how confidently does it assign a different label when there is a causal change in the input. Specifically, we define $\delta_{s} \in [0, 1]$ as:

\begin{align*}
    \delta_{s} &= \frac{(p(\hat{l}' | x') - p(\hat{l}' | x)) + (p(\hat{l} | x) - p(\hat{l} | x'))}{2} \text{,} 
\end{align*}
where $x = (P,H)$ is the original input and $x'$ is its perturbation.
Intuitively, this metric quantifies the amount of shift in model predictions between the two related examples. Unchanged model prediction results in a sensitivity of 0. When model prediction changes with extremely high confidence (i.e., assigning 100\% on its predicted labels), $\delta_s$ is normalized to 1. 
In binary classification, when the predicted label changes, the metric simplifies to:
\begin{align*}
    \delta_{s} &= p(\hat{l}' | x') + p(\hat{l} | x) - 1\text{.} 
\end{align*}
$\delta_{s}$ here measures the model's confidence in prediction when the context changes, shown by the probability it assigns to the predicted labels. In general, the higher the $\delta_{s}$, the more sensitive the model is to context changes in the input.

Next, we measure the counterfactual accuracy $\mathrm{Acc}_{\counterfactual}$. Under this metric, a prediction is correct only when the model correctly predicts the original and counterfactual examples. We use counterfactual accuracy to measure the consistency of model performance on original and counterfactual examples. $\mathrm{Acc}_{\counterfactual}$ is defined as:

{\footnotesize
\begin{align*}
     \frac{1}{K} \sum^{K}_{k=1} \mathbbm{1} \bigg((\hat{l}_k | \mathrm{P}_k,\mathrm{H}_k) = l^*_k) \land (\hat{l}'_k | \mathrm{P}'_k,\mathrm{H}_k) = l'^*_k)\bigg),
\end{align*}} where $K$ is the number of examples evaluated, $\hat{l}\text{,} \hat{l}'$ are model predictions for the original and counterfactual examples, and $l^*$, $l'^*$ are the gold labels, respectively.
\noindent This is similar in spirit to evaluations based on \emph{contrast sets} from \citet{gardner2020evaluating}, \emph{perturbation clusters} from \citet{khashabi2020more}, and the \emph{grouped probe metric} of \citet{trivedi2020dire}.

\begin{table}[t!]
\begin{center}
\resizebox{\columnwidth}{!}{
    \begin{tabular}{lcccccc}
      \toprule
       &  \multicolumn{2}{c}{\bf Human-CAD} & \multicolumn{2}{c}{\bf SNLI-hard$_{\counterfactual}$} & \multicolumn{2}{c}{\bf WANLI$_{\counterfactual}$} \\
      \cmidrule(lr){2-3} \cmidrule(lr){4-5} \cmidrule(lr){6-7}
      
      {\bf Method} & \multicolumn{1}{l}{\bf{ $\delta_{s}$}} & \multicolumn{1}{l}{\bf Acc$_{\counterfactual}$} & \multicolumn{1}{l}{\bf{\bf  $\delta_{s}$}} & \multicolumn{1}{l}{\bf Acc$_{\counterfactual}$}& \multicolumn{1}{l}{\bf $\delta_{s}$} & \multicolumn{1}{l}{\bf  Acc$_{\counterfactual}$} \\
      \midrule
       
      {SNLI-subset} & 62.8 & 59.1 & 66.1 & 51.1 & 51.3 & 39.3 \\
      + Tailor & 58.8 & 55.6 & 60.6 & 55.6 & 33.9 & 23.7 \\
      + Human-CAD & \bf{70.9} & 63.6 & 73.6 & 54.1 & 34.6 & 42.8 \\
      + \disco {\footnotesize (ours)} & 69.4 & 64.1 & \bf{74.3} & 60.3 & \bf{55.9} & 47.7 \\ \midrule

      {WANLI} & 41.4 & 30.5 & 47.4 & 27.0 & 44.5 & 42.1 \\
      + \disco {\footnotesize (ours)} & 65.6 & 64.9 & 68.5 & 59.2 & 46.1 & 42.8 \\ \midrule
      
      \disco {\footnotesize (ours)} & 65.7 & \bf{66.5} & 71.2 & \bf{63.1} & 41.9 & \bf{48.3} \\
      \bottomrule
    \end{tabular}
}
\caption{Performance on the counterfactual sensitivity tests including three datasets: Human-CAD, SNLI-hard$_{\counterfactual}$, AND WANLI$_{\counterfactual}$. The models used for evaluation are from the first experiment directly.} \label{tab:sensitivity}
\end{center}
\end{table}

\paragraph{Results} Table \ref{tab:sensitivity} shows models' performance on the three counterfactual evaluation sets. Models augmented or trained with \disco consistently outperform the baseline models by a large margin. Training with only \disco achieves the highest counter accuracy while augmenting \disco on the SNLI subset achieves the highest counterfactual sensitivity. This shows that our data helps increase the model's ability to differentiate the two examples and improve its reasoning performance on counterfactual data. Compared to other data augmentation methods, \disco yields a performance gain on both metrics showing its benefit on counterfactual reasoning. 

\disco increases the WANLI baseline's sensitivity and accuracy by more than 20\% and 30\% respectively on both Human-CAD and SNLI-hard$_{\counterfactual}$. However, the increase on WANLI$_{\counterfactual}$ is marginal, which is likely because \disco and the WANLI train set have very different distributions (OTDD distance 579). Although WANLI$_{\counterfactual}$ is close to the WANLI train set (OTDD distance 270), training on it yields lower accuracy than \disco, indicating that human-GPT-3 collaborated data construction does not necessarily grant models the ability to reason on counterfactual data. Thus, we can confirm that the distillation step on top of GPT-3 generation is essential for improving the model's counterfactual reasoning ability.

\section{Conclusion}
In this paper, we introduced the \disco framework for distilling high-quality counterfactual data from large language models (LLMs) using a task-specific teacher model for NLI. Through automatic and human evaluations, we show that counterfactuals generated by LLMs have higher quality and accuracy than human-written examples while having more diverse perturbations. Our evaluation results suggest that training or augmenting with distilled counterfactual data can help mitigate various types of distinct spurious patterns. Counterfactual examples produced by \disco significantly benefit model performance with improved robustness and out-of-distribution (OOD) generalizability. Despite a smaller data size, \disco data help models achieve better performance on the evaluation sets than baselines with extensive data. Furthermore, training on \disco examples improves model performance on counterfactual accuracy and helps the model be more sensitive to the context changes between counterfactual and original examples.

For future work, our method suggests several directions. While our efforts are limited to NLI, generating counterfactual data using LLMs is more general and, we believe, can be fruitfully applied to a wider range of tasks. In specific, only a task-specific filter model and modification to LLM prompts are needed to extend our generation pipeline to other tasks or even other languages. Also, while our approach takes inspiration from knowledge distillation \cite{hinton2015distilling} approaches and relies on a \emph{teacher} filtering model, alternative strategies could be used to improve the quality. As a related direction, techniques for semi-supervised learning over unfiltered LLM output should also be investigated to help utilize the wide range of data produced by LLMs.

\section{Limitations}

While we have argued that our approach to collecting counterfactual data via \disco is agnostic to the particular task and language, we emphasize that the experiments we report are limited to English and the task of NLI. Given that English is a high-resource language, there could be additional challenges (e.g., finding the tools needed for making span selection) in re-creating our pipeline for other languages. We also emphasize that our data generation experiments were carried out using only a single LLM, namely the publicly available GPT3 model first reported in \citet{brown2020language}.

As with the related studies we cite (e.g., \citet{liu2022wanli}), given the high costs associated with large-scale prompting, we are unable to ablate all parts of our data generation pipeline (e.g., the effect of systematically alternating prompting styles at scale, alternative span extraction techniques). Similar to virtually all experiments involving LLM prompting, such differences could affect the results and quality of the resulting augmentation datasets. Similarly, given the high costs of human annotation, we have limited our human evaluation to around 500 random instances (each involving 3 annotators), which follows other related studies. 

\section*{Acknowledgements}
We thank the anonymous reviewers for their
constructive and thoughtful comments. We also thank the members of the Aristo team at AI2 for providing helpful feedback on earlier versions of this work. Thanks finally to the Beaker team  (\url{https://www.beaker.org}) at AI2 for their assistance and help with experiments and computing infrastructure.

\bibliography{anthology,custom}
\bibliographystyle{acl_natbib}

\clearpage
\appendix

\section{Hyper-parameters and Implementation} 
\label{sec:hypers}

\paragraph{GPT3 and Teacher Model} For perturbation overgeneration, we use GPT-3 with the text-DaVinci-002 version. We set the {\it temperature} to 0.8 to encourage creative generations. For the penalties, we set the {\it frequency penalty} and {\it presence penalty} to 0.8 to lower the likelihood of sampling repeated words. To mitigate error propagation from the filtering step, we use a publicly available DeBERTa-v2 \cite{deberta} model checkpoint (containing 1.3 billion parameters) trained on a mixture of NLI datasets, including SNLI \cite{bowman2015large}, MultiNLI \cite{williams2017broad}, FEVER \cite{thorne2018fever}, ANLI \cite{anli}, that achieves SOTA performance on these datasets.

\paragraph{Student Models and Training Protocol} For all experiments, we tuned Robert-large (containing 345 million parameters) via a random search over key hyper-parameters in the style of \citet{devlin2018bert}. We used ADAM \cite{kingma2014adam} as our optimizer. The key hyper-parameters include \emph{learning rate} (including  $\mathtt{2e-5,3e-5,5e-5}$), \emph{batch size} (between $\mathtt{32,64}$), \emph{warmup ratio} (in the range of $\mathtt{0.08,0.1}$) and \emph{number of epochs} ($\mathtt{3}$ to $\mathtt{5}$); weight decay was kept constant at $0.1$ following \citet{liu2022wanli}, and early stopping was used with a patience of $2$ epochs. We generally found the following configuration to yield good performance: LR=$\mathtt{3e-5}$, epochs=$3$, batch\_size=$64$, warmup\_ration=$0.1$.  Standardly, model selection was performed by choosing the model with the highest validation accuracy. In our main result tables (i.e., Tables~\ref{tab:aug-results}-\ref{tab:gain}) we report the best of 5 models based on random restarts with different random seeds in all rows excluding the first 3. In the first 3 rows, given the large size of the training sets and the generally high cost of fine-tuning, we report the best single run (and generally found these models to yield low variance across hyper-parameters).

When comparing against other data augmentation approaches, e.g., Z-aug \cite{wu2022generating}, we used the exact code base compared with models trained on \disco to remove any differences in implementation (our implementation is based on the transformers library \cite{wolf2020transformers}).  All experiments were performed on an NVIDIA RTX A6000 GPU.

\section{Human Annotation Details}
\label{sec:amt}
\begin{figure*}
    \centering
    \includegraphics[width=0.95\textwidth]{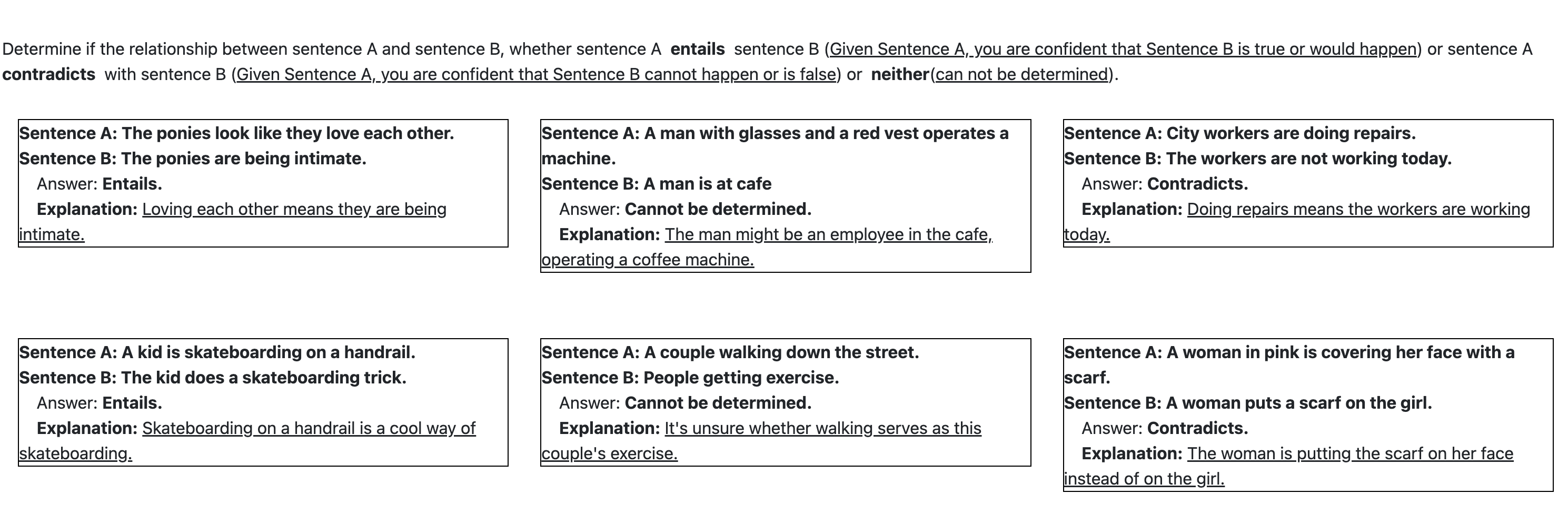}
    \caption{The annotated examples with explanations used on Amazon Mechanical Turk.}
    \label{fig:example_annot}
\end{figure*}

\begin{figure*}
    \centering
    \includegraphics[width=0.95\textwidth]{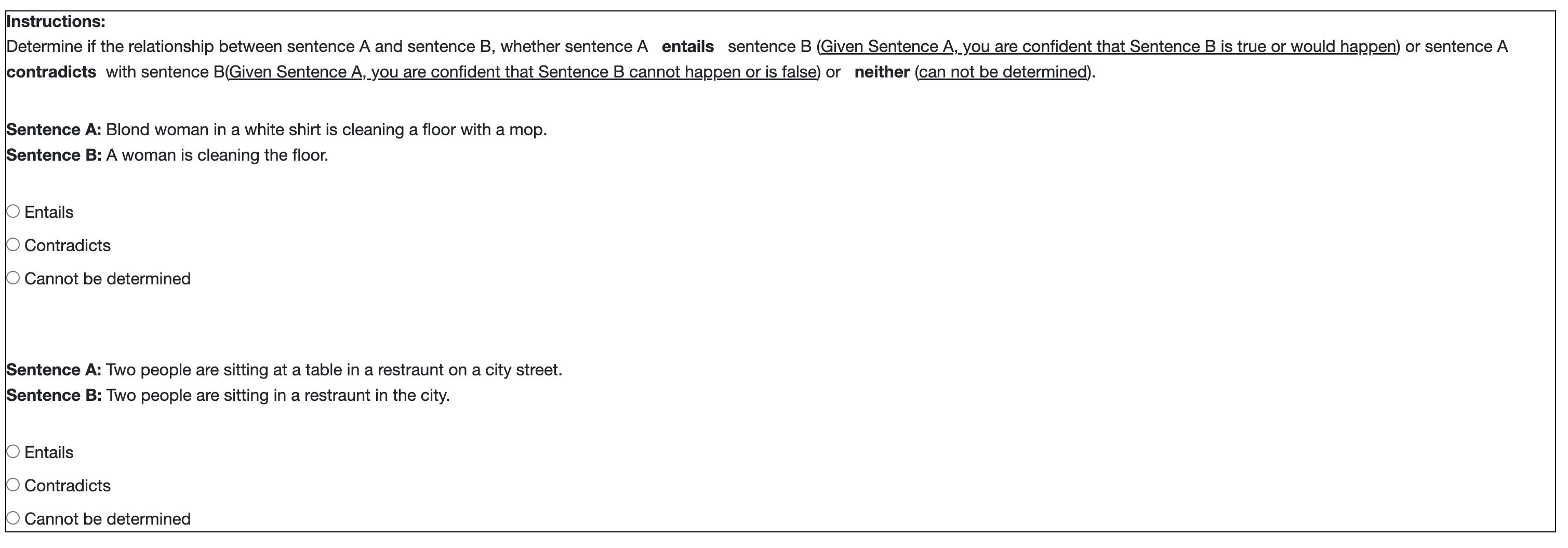}
    \caption{Instructions provided to human annotators on Amazon Mechanical Turk and the annotation interface.}
    \label{fig:instruction}
\end{figure*}

We recruit human annotators to evaluate our generated counterfactual data and to annotate two evaluation sets for counterfactual consistency: SNLI-hard$_{\counterfactual}$ and WANLI$_{\counterfactual}$. Here we discuss the details of our annotation studies. Screenshots of the instructions, guidelines, and annotation interface are shown in Fig \ref{fig:example_annot} and Fig \ref{fig:instruction}.

\paragraph{Annotators} We recruit human workers on the Amazon Mechanical Turk \footnote{\url{https://www.mturk.com/}} platform. We required Mechanical Turk Masters to perform our tasks. Annotators must have a HIT approval rate of 98\%, a total of 1000 approved HITs, and be in the United States. Throughout the data collection process, we randomly select a subset of the annotations to check and correct any potentially controversial annotations. For each problem, we assign three annotators and use a majority vote to determine the final annotation. Workers were paid \$0.3 for each AMT hit (consisting of 10 examples to annotate).

\end{document}